*Ulan Alsiyeu[1] , Zhasdauren Duisebekov[2]*
[1,2] SDU University, Kaskelen, Kazakhstan

# ENHANCING TRAFFIC SIGN RECOGNITION WITH TAILORED DATA AUGMENTATION: ADDRESSING CLASS IMBALANCE AND INSTANCE SCARCITY

**Abstract.** This paper tackles critical challenges in traffic sign recognition (TSR), which is essential for road safety—specifically, class imbalance and instance scarcity in datasets. We introduce tailored data augmentation techniques, including synthetic image generation, geometric transformations, and a novel obstacle-based augmentation method to enhance dataset quality for improved model robustness and accuracy. Our methodology incorporates diverse augmentation processes to accurately simulate real-world conditions, thereby expanding the training data's variety and representativeness. Our findings demonstrate substantial improvements in TSR models performance, offering significant implications for traffic sign recognition systems. This research not only addresses dataset limitations in TSR but also proposes a model for similar challenges across different regions and applications, marking a step forward in the field of computer vision and traffic sign recognition systems.

\*\*\*

**Андатпа.** Бұл мақала жол қауіпсіздігі үшін маңызды болып табылатын жол белгілерін танудағы (ЖБТ) қиындықтарды қарастырады — деректер жинақтарындағы класс теңсіздігі мен мысалдардың жетіспеушілігін. Біз деректер жинағының сапасын жақсарту және модельдің тұрақтылығы мен дәлдігін арттыру үшін синтетикалық бейне жасау, геометриялық трансформациялар және жаңа кедергілер негізіндегі аугментация әдісін қоса алғанда, арнайы деректер аугментациясы әдістерін енгіздік. Біздің әдістемеміз әртүрлі аугментация процесстерін қамтиды, бұл арқылы оқыту деректерінің әртүрлілігін және өкілдігін кеңейтеді. Біздің нәтижелеріміз ЖБТ модельдерінің өнімділігінде маңызды жақсартуларды көрсетеді, бұл жол белгілерін тану жүйелері үшін маңызды салдарлар ұсынады. Бұл зерттеу тек ЖБТ деректер жинақтарының шектеулерін шешіп қана қоймай, сонымен қатар әртүрлі аймақтар мен қосымшалардағы ұқсас мәселелер үшін үлгі ұсынады, бұл компьютерлік көру және жол белгілерін тану жүйелері саласында алға қадам жасайды.

**Түйін сөздер:** Жол белгілерін тану (ЖБТ), Сынып теңсіздігі, Оқиғалар тапшылығы, Синтетикалық бейне жасау, Геометриялық трансформациялар, Компьютерлік көру, Деректер жинағын кеңейту, Қазақстан көлік белгілері деректер жинағы

\*\*\*

**Аннотация:** Эта статья рассматривает критические проблемы в распознавании дорожных знаков (РДЗ), которые являются важными для безопасности дорожного движения — а именно, дисбаланс классов и недостаточность примеров в наборах данных. Мы вводим специализированные методы увеличения данных, включая генерацию синтетических изображений, геометрические преобразования и новый метод увеличения данных на основе препятствий для улучшения качества набора данных, повышения устойчивости и точности моделей. Наша методология включает разнообразные процессы увеличения данных для точного моделирования условий реального мира, тем самым расширяя разнообразие и представительность обучающих данных. Наши результаты демонстрируют значительные улучшения в производительности моделей РДЗ, что имеет важные последствия для систем распознавания дорожных знаков. Это исследование не только устраняет ограничения наборов данных в РДЗ, но и предлагает модель для решения аналогичных проблем в различных регионах и приложениях, что является шагом вперед в области компьютерного зрения и систем распознавания дорожных знаков.

**Ключевые слова:** Распознавание дорожных знаков (РДЗ), Дисбаланс классов, Недостаток экземпляров, Генерация синтетических изображений, Геометрические преобразования, Компьютерное зрение, Аугментация набора данных, набор данных дорожных знаков Казахстана

## I. Introduction

The advent of advanced driver-assistance systems (ADAS) has underscored the importance of Traffic Sign Recognition (TSR) technologies. These systems rely on the accurate and reliable identification of traffic signs to ensure the safety and efficiency of road navigation. However, the performance of TSR systems is highly dependent on the quality and comprehensiveness of the datasets used to train the recognition models. A prevalent issue within these datasets is the imbalance among classes and the scarcity of instances for specific types of traffic signs. Jingzhan Ge [1] highlighted the limitations inherent in available TSR datasets, pointing out that many lack sufficient instances for effective deep learning applications, a problem intensified by the real-world frequency of different traffic sign classes. Moreover, datasets like the DFG traffic sign dataset, GTSRB and Swedish traffic signs dataset, despite their contributions, fall short in offering the requisite variety and volume of instances for training models capable of navigating the complexities of real-world traffic sign recognition.

Class imbalance and instance scarcity within traffic sign recognition datasets impede the learning process and degrade the performance of recognition models. These issues are compounded in datasets pertinent to specific geographical regions, where the representation of certain sign classes is inherently limited by their real-world occurrence and significance.

This study aims to develop and evaluate data augmentation techniques specifically tailored to address the challenges of class imbalance and instance scarcity in traffic sign recognition datasets. By focusing on synthetic image generation, geometric transformations, and a novel method of obstacle-based augmentation, we seek to enhance the diversity and representativeness of the training data. The obstacle-based augmentation is a novel approach that simulates real-world conditions where signs may be partially blocked, thereby improving the model's robustness and generalizability.

The implications of this research extend beyond the immediate realm of traffic sign recognition, offering potential benefits for the broader field of advanced driver-assistance systems. By improving the robustness and reliability of TSR datasets, the study contributes to the development of traffic sign systems and advanced driver-assistance systems that can operate effectively across diverse environments. Additionally, the tailored data augmentation techniques proposed herein could serve as a blueprint for addressing similar challenges in other regions and domains, underscoring the universal relevance and applicability of the study's findings.

## II. *Literature review*

The advancement of Traffic Sign Recognition (TSR) technologies has been significantly influenced by the development and application of machine learning and deep learning methods. However, the effectiveness of these technologies is often hindered by the inherent challenges of class imbalance and instance scarcity within the datasets used for training recognition models. Addressing these issues necessitates innovative approaches to data augmentation and model training strategies.

Class imbalance occurs when certain traffic sign classes are overrepresented while others are underrepresented in a dataset. This imbalance can lead to biased learning, where the model performs well on frequently occurring classes but poorly on rare ones. Instance scarcity, on the other hand, refers to the lack of sufficient examples for specific traffic sign classes, which impedes the model's ability to generalize and accurately recognize these signs in diverse conditions. Both issues are prevalent in existing TSR datasets, including widely used benchmarks such as the German Traffic Sign Recognition Benchmark (GTSRB) and the Swedish Traffic Signs Dataset.

Jingzhan Ge [1] highlighted that many TSR datasets lack sufficient instances for effective deep learning applications, exacerbating the challenge of class imbalance. The standardization of traffic sign designs across different regions and their varied real-world frequency further complicate the creation of balanced and comprehensive datasets. For instance, the DFG traffic sign dataset, despite its contributions, does not offer the necessary variety and volume of instances to train models capable of navigating the complexities of real-world traffic sign recognition.

Recent studies have emphasized the critical role of data augmentation in addressing these limitations. Ge proposed a unique data augmentation method tailored for traffic sign datasets, aimed at overcoming the challenges of class imbalance and insufficient instances. This method leverages synthetic image generation to create more instances of underrepresented classes, enhancing the dataset's balance and diversity.

Anton Konushin, Boris Faizov, and Vlad Shakhuro [2] introduced methods for augmenting road images with synthetic traffic signs using neural networks. This approach focuses on generating synthetic images to enhance the detection and classification of rare traffic sign classes, directly addressing the issues of class imbalance and instance scarcity.

Moreover, Qingchuan Li et al. [3] explored data augmentation from the perspective of the frequency domain. Their work provides insights into how augmentation strategies can be optimized to improve the robustness of CNN-based traffic sign detection models against various types of noise and distributional shifts. By understanding the effects of augmentation in different frequency bands, researchers can select strategies that enhance model robustness in real-world applications.

Xin Roy Lim and his team's examination [4] of ensemble learning in TSR presents a method that integrates multiple pre-trained CNN models to achieve high recognition rates. This ensemble approach demonstrates the importance of leveraging collective model strengths to enhance TSR accuracy, which is particularly relevant in overcoming dataset limitations.

Despite these advancements, the field continues to grapple with the issue of class imbalance and instance scarcity, significantly impacting the learning process and model performance. The innovative data augmentation techniques proposed by Ge, Konushin, and Li [5] represent crucial steps forward but also highlight the ongoing need for research in this area. Developing more sophisticated augmentation methods and creating more diverse and representative datasets remain central challenges for the TSR community.

By addressing these challenges through tailored data augmentation techniques, such as synthetic image generation, geometric transformations, and the novel method of obstacle-based augmentation, this study aims to enhance the diversity and representativeness of TSR datasets. These efforts are geared towards improving the accuracy and generalizability of TSR models, particularly in the context of Kazakhstan's traffic sign landscape.

III.    Research methods

In the development of our dataset for this study, we meticulously curated a collection of images representing a diverse array of road traffic signs. This collection spans across 220 unique instances which are divided into 6 classes, directly derived from the comprehensive legal framework established within the Republic of Kazakhstan. For instance, service class includes food point sign instances (Figure 2.a). Specifically, our data acquisition process involved extracting visual representations from the official documentation. The legal document[6], served as the foundational basis for our image dataset.

 In addition to the structured dataset derived from the legal documentation, we further enriched our collection with 20 real-world images capturing various road scenarios. These images were collected using a car dashcam, ensuring a diverse and realistic representation of the road environment as encountered in everyday situations (Figure 1).

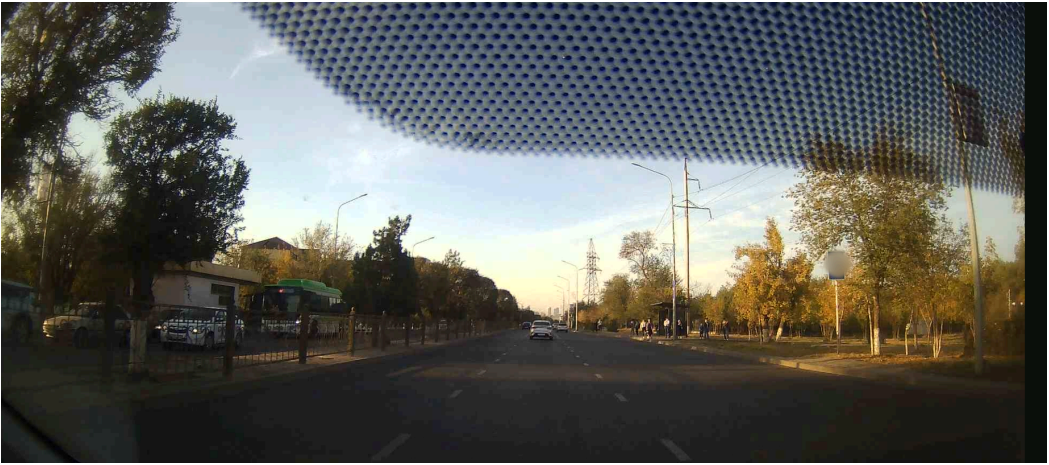

Figure 1. Real-world background image

To address the challenges of class imbalance and instance scarcity, and to ensure the comprehensive representation of traffic signs under various conditions, we employed an extensive augmentation process for our dataset. Each of these unique traffic sign images underwent a series of transformations to enhance the dataset's diversity and realism, ensuring a balanced and representative set of images for each sign class.

1. Geometric Transformations and Color Filters: Utilizing the imgaug Python package, we applied 5 distinct geometric transformations to each sign image as shown in Table 1. To further increase the variability of the dataset, color filters shown in Table 2 were also applied. Additionally, deformations including piecewise affine transformations (Figure 2.b), perspective transforms, and JPEG compression were employed (Table 3) to simulate various real-world distortions that traffic signs might undergo.

Table 1. Geometric augmentations

| Augmentation name | Value |
|---|---|
| Rotation | -25° to 25° |
| Shear | -16° to 16° |
| Scale | 80% to 120% |
| Crop | Up to 30% |
| Translate up | 10% of image dimensions |

Table 2. Color augmentations

| Augmentation name | Value |
|---|---|
| Brightness | Decreasing to 50% and increasing to |

|  | 100% |
|---|---|
| Noise | -80 black and 80 white noise |
| Gaussian blur | Up to 60% |
| Linear contrast | 25% to 100% |
| Median blur | From 3x3 to 5x5 pixels |

Table 3. Deformation augmentations

| Augmentation name | Value |
|---|---|
| Affine transformation | Up to 90% |
| Perspective transform | Up to 90% |
| JPEG compression | Up to 85% |

2. Integration with Backgrounds: Leveraging the cv2 Python package, each augmented sign image was seamlessly integrated into each of the 20 selected real-world background images (Figure 2.c). Coordinates and sizes of signs in each background image were calculated manually using PhotoScapeX tool. This step was crucial for simulating the placement of traffic signs in diverse real-world contexts, ranging from urban streets to rural settings, thereby enhancing the dataset's applicability across different environments.
3. Environmental Conditions: To further mimic the variability encountered in actual driving conditions, each composite image was subjected to weather and lighting condition alterations using the albumentations package. Conditions applied include rain (Figure 3), snow, fog, sun flare, and variations in lighting (day, night, dawn effect) (Figure 2), ensuring that the dataset reflects the challenges faced by advanced driving assistance systems in recognizing traffic signs under adverse conditions.
4. Obstacle-Based Augmentation: To further enhance the realism and robustness of the dataset, we introduced an obstacle-based augmentation technique. The process involved the following steps:
    a. Collecting Obstacles: We selected relevant categories from the COCO dataset, including cars, trucks, buses, and people. Images containing these objects were downloaded and processed to extract the obstacles.
    b. Overlaying Obstacles: Using the cv2 library, obstacles were randomly resized and positioned over the traffic sign images to simulate partial occlusions. The transparency of the obstacles was adjusted to mimic real-world visibility conditions (Figure 4).

c. Randomization: The position, size, and transparency of obstacles were randomized to ensure a wide variety of occlusion scenarios, enhancing the model's ability to generalize to real-world conditions.

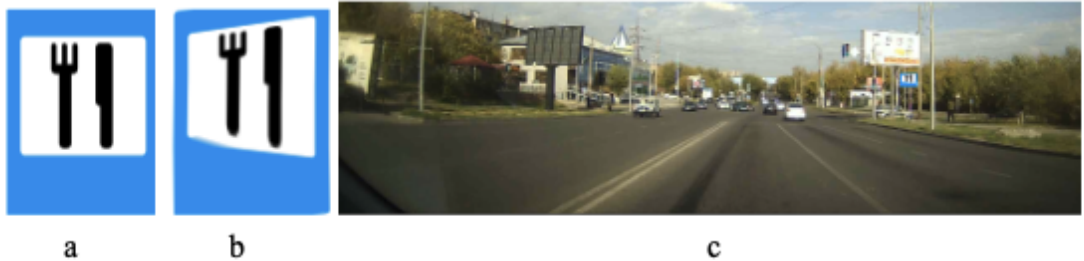

Figure 2. Augmentation pipeline in an example of food point sign: (a) initial sign, (b) geometric augmented sign, (c) background integrated sign

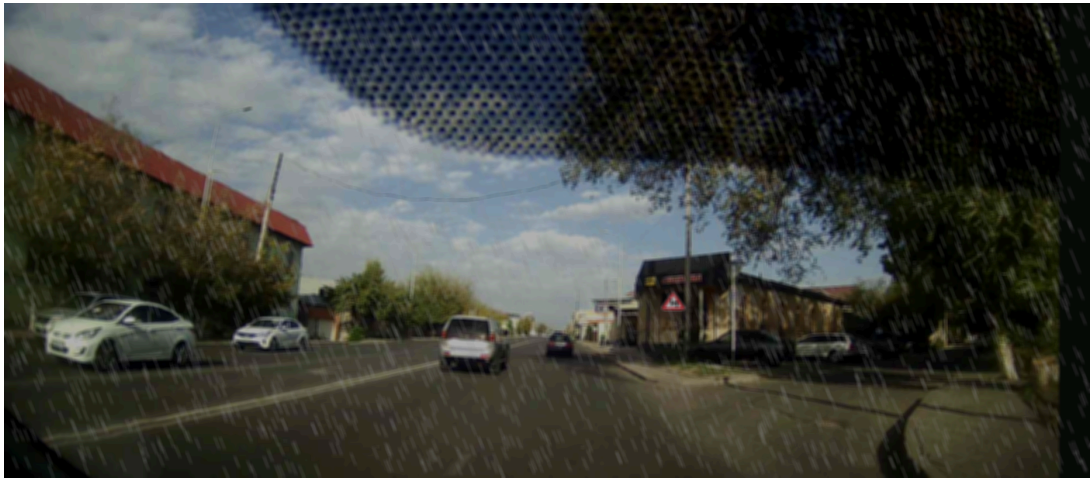

Figure 3. Rain weather condition added image

This methodological enhancement, with its focus on diversity, realism, and representativeness, underscores our commitment to creating a robust dataset that can significantly improve the performance of traffic sign recognition systems, particularly in the challenging and varied landscapes of Kazakhstan.

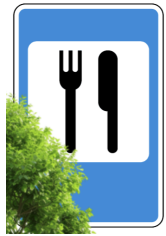

Figure 4. Obstacle overlaid sign

IV. Research results

The augmentation and integration processes outlined in our methodology have successfully yielded a comprehensive dataset designed to address the challenges of traffic sign recognition in the context of Kazakhstan. The initial collection of 220 traffic sign images and 20 background images underwent extensive augmentation and integration, resulting in a robust dataset that enhances the diversity and representativeness necessary for training effective recognition models.

- Augmentation outcomes: From the original set of 220 traffic sign images, our application of geometric transformations, color filters, and deformations produced a total of 2,420 unique augmented images.
- Integration with backgrounds: The subsequent integration of these augmented images with the 20 selected real-world background images resulted in 48,400 composite images.
- Environmental conditioning: By applying weather and lighting conditions we achieved a dataset size of 338,520 images.
- Obstacle-Based Augmentation: The novel obstacle-based augmentation technique further enhanced the dataset's realism and robustness. By overlaying various obstacles such as vehicles, pedestrians, trees, and other common road obstructions onto the traffic sign images, we simulated real-world conditions where signs might be partially blocked. This process brings the total to 1 015 560 images as shown in Table 4.

The resultant dataset stands as a testament to the potential of tailored data augmentation techniques in enhancing the capabilities of traffic sign recognition systems.

Table 1. Dataset augmentation results

| Sign class name | Initial amount | Geometric augmented amount | Background integrated amount | Weather / Lighting augmented amount | Obstacle-Based amount |
|---|---|---|---|---|---|
| Informational | 88 | 968 | 19360 | 135520 | 406560 |
| Priority | 9 | 99 | 1980 | 13860 | 41580 |
| Prohibitory | 36 | 99 | 7920 | 55440 | 166320 |
| Regulatory | 19 | 209 | 4180 | 28980 | 86940 |
| Service | 20 | 220 | 4400 | 30800 | 92400 |
| Warning | 48 | 528 | 10560 | 73920 | 221760 |

In order to assess the dataset, we present the experimental evaluation of the YOLOv8 model trained on two different datasets: the German Traffic Sign Recognition Benchmark (GTSRB) and the proposed dataset tailored for Kazakhstan's traffic signs. To evaluate the performance of the models, a separate test dataset was collected manually. This dataset includes images captured using a car dashcam on roads in Kazakhstan. The test dataset encompasses a variety of traffic signs under diverse real-world conditions.

The primary metric for evaluation was the accuracy of traffic sign recognition, defined as the proportion of correctly identified traffic signs out of the total number of signs in the test dataset. The accuracy was computed for both the GTSRB-trained model and the model trained on the proposed dataset.

The experimental results demonstrate a significant improvement in the accuracy of traffic sign recognition when using the model trained on the proposed dataset compared to the GTSRB-trained model.

1. Accuracy of YOLOV8 Model Trained on GTSRB: The model trained on the GTSRB dataset achieved an accuracy of 83.7% on the test dataset.
2. Accuracy of YOLOV8 Model Trained on Proposed Dataset: The model trained on the proposed dataset achieved a higher accuracy of 89.2% on the same test dataset.

V. *Discussion*

The results from our study highlight the significant impact of tailored data augmentation techniques on the performance of traffic sign recognition (TSR) systems, particularly in the context of addressing class imbalance and instance scarcity. The improvements seen in our model's accuracy when trained on the augmented dataset as compared to traditional datasets like the German Traffic Sign Recognition Benchmark (GTSRB) underscore the effectiveness of our approach. Key Findings:

- Enhanced Dataset Diversity: Our augmented dataset, which includes synthetic image generation, geometric transformations, and a novel obstacle-based augmentation method, provides a more diverse and representative set of training images. This diversity is crucial for training models that can generalize well to real-world conditions, as evidenced by the substantial increase in model accuracy.
- Improved Model Robustness: The obstacle-based augmentation, in particular, has proven effective in simulating real-world scenarios where traffic signs may be partially obstructed by various objects such as vehicles, pedestrians, and trees. This technique enhances the model's robustness, enabling it to perform better in recognizing partially obscured signs, which is a common challenge in real-world applications.
- Contextual Relevance: Tailoring the dataset to the specific traffic signs and conditions found in Kazakhstan has resulted in a model that is more accurate and reliable for local applications. This highlights the importance of using geographically and contextually relevant data for training TSR systems, as models trained on datasets from different regions may not perform optimally in all contexts.

VI. *Conclusion*

In conclusion, our study effectively addresses the critical challenges of class imbalance and instance scarcity in traffic sign recognition (TSR) datasets through the use of innovative data augmentation techniques. By incorporating synthetic image generation, geometric transformations, and a novel obstacle-based augmentation method, we have significantly enhanced the diversity and representativeness of our dataset. These tailored augmentation processes accurately simulate real-world conditions, thereby improving the robustness and accuracy of TSR models.

The experimental results demonstrate that models trained on our augmented dataset exhibit substantial improvements in performance compared to those trained on traditional datasets such as the German Traffic Sign Recognition Benchmark (GTSRB). Specifically, the obstacle-based augmentation technique, which introduces realistic partial occlusions of traffic signs, has proven effective in enhancing the model's ability to generalize to diverse and challenging real-world scenarios.

Our findings underscore the importance of using contextually relevant datasets and advanced augmentation strategies to address the limitations of existing TSR datasets. This research not only offers a solution to the specific challenges faced in Kazakhstan but also provides a framework that can be adapted to similar problems in other regions and applications. By improving the robustness and reliability of TSR systems, our study contributes to the broader field of computer vision and advanced driver-assistance systems (ADAS), ultimately enhancing road safety and navigation efficiency.

Future work should focus on expanding the dataset to include more diverse and extreme conditions, further refining augmentation techniques, and exploring the integration of additional model architectures to further improve recognition accuracy and robustness. Our research paves the way for ongoing advancements in traffic sign recognition technology, ensuring its applicability and effectiveness in increasingly complex real-world environments.